\pdfoutput=1

\documentclass[11pt]{article}

\usepackage[table]{xcolor}
\usepackage{booktabs}
\usepackage{multirow}
\usepackage{array,graphicx}
\usepackage{listings}
\usepackage{adjustbox}

\usepackage{acl}

\usepackage{times}
\usepackage{latexsym}

\usepackage{amssymb}
\usepackage{amsmath}

\usepackage[T1]{fontenc}

\usepackage[utf8]{inputenc}

\usepackage{linguex}

\usepackage{graphicx}
\usepackage{subcaption}

\usepackage{microtype}

\title{MELA: Multilingual Evaluation of Linguistic Acceptability}

\author{
Ziyin Zhang$^\dagger$$^*$, Yikang Liu$^\ddagger$$^*$, Weifang Huang$^\ddagger$, Junyu Mao$^\diamond$, Rui Wang$^\dagger$, Hai Hu$^\ddagger$ \\
$^\dagger$ Dept.~of Computer Science and Engineering, Shanghai Jiao Tong University\\
$^\ddagger$ School of Foreign Languages, Shanghai Jiao Tong University\\
$^\diamond$ School of Arabic Studies, Beijing Foreign Studies University\\
{\footnotesize
\texttt{\{daenerystargaryen;yikangliu;huangweifang;wangrui12;hu.hai\}@sjtu.edu.cn}} \\
{\footnotesize \texttt{maojunyu@bfsu.edu.cn}}\\
  }

\begin{document}
\maketitle

\def\thefootnote{*}\footnotetext{First two authors contributed equally to this work. Corresponding authors: Rui Wang and Hai Hu.}\def\thefootnote{\arabic{footnote}}

\begin{abstract}
In this work, we present the largest benchmark to date on linguistic acceptability: Multilingual Evaluation of Linguistic Acceptability---MELA, with 46K samples covering 10 languages from a diverse set of language families. We establish LLM baselines on this benchmark, and investigate cross-lingual transfer in acceptability judgements with XLM-R. In pursuit of multilingual interpretability, we conduct probing experiments with fine-tuned XLM-R to explore the process of syntax capability acquisition. Our results show that GPT-4o exhibits a strong multilingual ability, outperforming fine-tuned XLM-R, while open-source multilingual models lag behind by a noticeable gap. Cross-lingual transfer experiments show that transfer in acceptability judgment is non-trivial: 500 Icelandic fine-tuning examples lead to 23 MCC performance in a completely unrelated language---Chinese.  
Results of our probing experiments indicate that training on MELA improves the performance of XLM-R on syntax-related tasks.

\includegraphics[width=1em,height=1em]{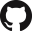}\hspace{.5em}\parbox{\dimexpr\linewidth-2\fboxsep-2\fboxrule}{\footnotesize \url{https://github.com/sjtu-compling/MELA}}
\end{abstract}

\section{Introduction}\label{sec:intro}

\begin{table*}[t]
\centering
\includegraphics[width=1\textwidth]{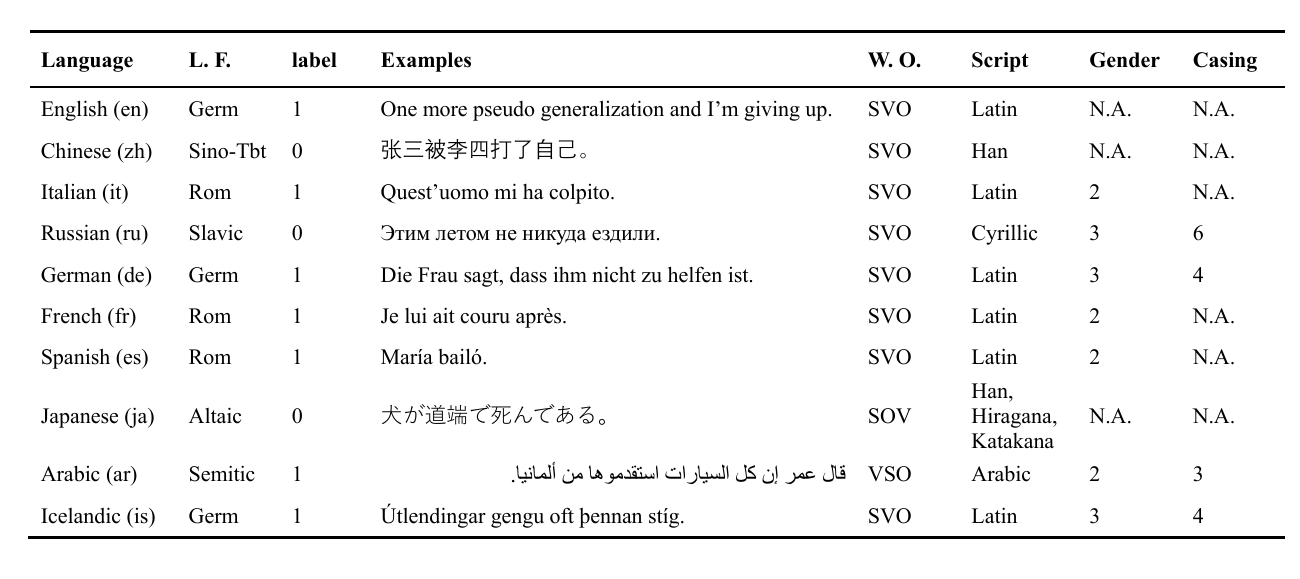}
\caption{Example sentences in the MELA training set, with information about the language family (L.F.), word order (W.O.), script, grammatical gender and casing for each language. Label ``1'' indicates the sentence is acceptable, ``0'' unacceptable.  Data for the first four languages are from existing benchmarks while the rest are collected by us.}
\label{tab:examples}
\end{table*}

The acceptability judgment task tests a language model's ability to distinguish syntactically acceptable sentences like (1a) from unacceptable ones like (1b) in a human language - for instance, the following example on island constraints in English~\citep{ross1967}.

\ex. \a. Whose book did you find?
     \b. *Whose did you find book?

As a core linguistic competence, the ability to tell well-formed sentences from ill-formed ones is one of the first that a good language model should have. 

Many corpora and benchmarks have been built to evaluate language models' syntactic ability, 
using either a data-driven approach, where examples created by theoretical linguists in published textbooks are collected, e.g., CoLA---Corpus of Linguistic Acceptability~\citep{2018CoLA}, or a theory-driven approach, where minimal pairs targeting specific syntactic phenomena are generated semi-automatically via some template 
\citep{2019BLiMP,2021CLiMP,hu-etal-2020-systematic}. 

Recently, there has been growing interest in expanding the data-driven paradigm into other languages. 
For instance, CoLA-style datasets have been proposed in Russian~\citep{2022RuCoLA}, Italian~\citep{2021ItaCoLA}, and Chinese~\citep{2023CoLAC}. 
However, to date there are almost no multilingual benchmarks in this area that can be used to systematically test such abilities of multilingual models. 

On the other hand, recently introduced benchmarks for Large Language Models (LLMs) such as GPT-4~\citep{2023GPT4} have mostly focused on application-driven tasks such as world knowledge and commonsense reasoning~\citep{2020MMLU,2022BIG-Bench}, math reasoning~\citep{2021GSM8K,2021MATH}, and code generation~\citep{2021Codex,2021MBPP,2023codesurvey}. Few studies, however, have investigated these models from a more linguistics-oriented aspect.

To address these gaps, we introduce MELA---Multilingual Evaluation of Linguistic Acceptability, the first large-scale multilingual acceptability benchmark with 46k examples covering 10 languages from a diverse set of language families. 
Data in four languages are from existing benchmarks mentioned above, 
and we 
complement them with newly collected data in six languages. Examples of MELA are demonstrated in Table~\ref{tab:examples}. Following the CoLA tradition, all sentences in MELA are hand-written by linguists in respective languages, taken from textbooks, handbooks and journal articles in theoretical syntax, except for a small fraction of Russian sentences from \citet{2022RuCoLA}. 

We propose three possible usage cases for MELA, and make  preliminary explorations in this paper:

\paragraph{Benchmarking} We benchmark various multilingual language models (LMs) on MELA, including BLOOMZ~\citep{2022BLOOM,2022xP3},  mTk~\citep{2022SuperNatural}, mT0~\citep{2022xP3}, Baichuan2-Chat~\citep{2023Baichuan2}, GPT-3.5 and GPT-4o~\citep{2023GPT4}.
\paragraph{Cross-lingual transfer} We train XLM-R~\citep{2019XLM-R} on different language combinations, finding non-trivial cross-lingual transfer performance even between unrelated language pairs, despite the vast difference in the basic syntax of the 10 languages in MELA. 
\paragraph{Syntax acquisition} We probe the syntactic capacity of MELA-finetuned XLM-Rs on syntax-related probing tasks, which indicates that XLM-R acquires some syntactic knowledge from finetuning on the acceptability judgment task. 

In the rest of this paper, we first review relevant literature in \S\ref{sec:related}, and then describe how MELA was constructed in \S\ref{sec:mela}. 
Next, we use MELA to benchmark several open-source and close-source LLMs in \S\ref{sec:llm}. We investigate cross-lingual transfer and multilingual fine-tuning in \S\ref{sec:transfer}.
Finally, we probe the XLM-Rs trained on MELA for their syntax-related capacity in \S\ref{sec:analysis}.

\section{Related Work}\label{sec:related}
\subsection{Linguistic Acceptability}
As we mentioned in \S\ref{sec:intro}, large-scale linguistic acceptability datasets are currently available for four languages: CoLA for English~\citep{2018CoLA}, ItaCoLA for Italian~\citep{2021ItaCoLA}, RuCoLA for Russian~\citep{2022RuCoLA}, CoLAC for Chinese~\citep{2023CoLAC}, NoCoLA for Norwegian~\citep{jentoft2023nocola}, and JCoLA~\citep{someya-etal-2024-jcola-japanese} for Japenese. Sentences from these datasets are taken from academic works by theoretical syntacticians and are therefore annotated by expert linguists.\footnote{CoLAC also comes with an additional set of crowd labels; Unacceptable sentences in NoCoLA are sourced from grammatical mistakes made by Norwegian learners.} 

Another line of work in linguistic acceptability is based on minimal pairs, consisting of two near-identical sentences with minimal differences. 
Language models are expected to assign a higher probability to the acceptable sentence than the unacceptable one.
The minimal pair paradigm is adopted to evaluate specific syntactic issues such as subject-verb agreement~\citep{linzen2016assessing,marvin2018targeted,crosslingualsyntax}, reflexive anaphora~\citep{futrell2019neural,hu2020closer}, negative polarity licensing~\citep{wilcox2019structural,jumelet2018language}, long-distance dependency~\citep{wilcox2018rnn,chowdhury2018rnn}, and argument structure~\citep{kann2019verb,tjuatja2023syntax}.
Following these works, comprehensive benchmarks of minimal pairs are constructed in English resources~\citep{2019BLiMP,hu-etal-2020-systematic}, and then expanded into other languages~\citep{2021CLiMP,song2022sling,someya2023jblimp,nielsen-2023-scandeval}.

In this work, we follow CoLA in constructing our benchmark as an initial step 
towards multilingual evaluation in acceptability judgment, as our goal 
is to have a wide coverage of syntactic phenomena in the languages selected.

\subsection{Multilingual Evaluation Benchmarks}
XTREME~\citep{2020XTREME} and XGLUE~\citep{2020XGLUE} are two of the most popular multilingual evaluation benchmarks. Of the tasks therein, many are constructed by translating English samples entirely or partially into other languages, such as XNLI~\citep{2018XNLI}, PAWS-X~\citep{2019PAWS-X}, and MLQA~\citep{2019MLQA}.

Apart from these NLU benchmarks, the literature has also witnessed an abundance of multilingual generation benchmarks, ranging from summarization~\citep{2020MLSUM,2020WikiLingua} to translation~\citep{2020M2M,2021Flores101}. After the popularization of multitask instruction finetuning in language models~\citep{2021FLAN,2021T0}, multilingual instruction datasets have also been proposed, represented by Supernatural Instruction~\citep{2022SuperNatural} and xP3~\citep{2022xP3}. We refer to \citet{2024resourcesurvey} for a more comprehensive review of recent multilingual resources.

\begin{table*}[ht]
    \centering
    \adjustbox{width=\textwidth,center}{
    \begin{tabular}{lcccccccccc}
    \toprule
        & English & Chinese & Italian & Russian & German & French & Spanish & Japanese & Arabic & Icelandic\\
        ISO code & en & zh & it & ru & de & fr & es & ja & ar & is\\
    \midrule
Train$_{v1.0}$ & 8551 & 6072 & 7801 & 7869 & 500 & 500 & 500 & 500 & 500 & 500 \\
Dev$_{v1.0}$ & 527 & 492 & 946 & 1405 & 272 & 466 & 295 & 580 & 258 & 899 \\
Test$_{v1.0}$ & 516 & 931 & 975 & 2227 & 273 & 467 & 293 & 581 & 259 & 899 \\\midrule
Train$_{v1.1}$ & 8551 & 6072 & 7801 & 7869 & - & - & - & - & - & - \\
Dev$_{v1.1}$ & 527 & 492 & 946 & 1405 & 100 & 100 & 100 & 100 & 100 & 100 \\
Test$_{v1.1}$ & 516 & 931 & 975 & 2227 & 945 & 1333 & 988 & 1561 & 917 & 2198 \\\midrule
acceptable%
len (char) & 33.1 & 10.7 & 30.0 & 47.9 & 39.9 & 22.9 & 26.2 & 14.7 & 18.1 & 26.4 \\
len (byte) & 34.1 & 34.3 & 31.3 & 95.7 & 41.5 & 24.6 & 28.4 & 47.0 & 38.3 & 31.1 \\
len (token) & 10.5 & 9.5 & 9.7 & 15.3 & 11.4 & 8.1 & 8.7 & 10.9 & 8.2 & 9.4 \\\bottomrule
    \end{tabular}
    }
    \caption{Statistics of MELA: train/dev/test splits (in number of sentences), acceptable rate, and average sentence length by characters, bytes, and tokens (using the tokenizer of XLM-R~\citep{2019XLM-R}). Subscripts denote the version of data splits: \textit{v}1.0 is used for XLM-R fine-tuning and \textit{v}1.1 is used for LLM zero/few-shot experiments.
    }
    \label{tab:mela}
\end{table*}

\section{MELA: Multilingual Evaluation of Linguistic Acceptability}\label{sec:mela}
MELA consists of more than 46 thousand acceptability samples across 10 languages from a diversity of language families and groups. Specifically, it contains three Germanic languages:
English, German and Icelandic, three
Romance languages: Spanish, French and Italian, one Slavic language: Russian, one  Sino-Tibetan language: Chinese, one Japonic language: Japanese, and one Semitic language: Arabic. Table~\ref{tab:examples} shows example sentences and properties of each language in MELA. For dataset statistics, see Table~\ref{tab:mela}.

\subsection{Data collection Procedure}

\paragraph{High-resource languages.} 
We use four existing datasets for four languages in MELA: CoLA~\citep{2018CoLA} for English, ItaCoLA~\citep{2021ItaCoLA} for Italian, RuCoLA~\citep{2022RuCoLA} for Russian, and CoLAC for Chinese~\citep{2023CoLAC}, each having more than 6,000 data points.\footnote{NoCoLA and JCoLA are not included for they are concurrent with this work.}
Since the out-of-domain samples of RuCoLA are produced by generative models, we additionally collected 1037 Russian samples from \textit{The Syntax of Russian}~\citep{syntaxRussian} (with the procedure described below) and add them 50-50 to the development and test sets of the Russian portion to keep a balance between validation-test discrepancy and generalization.

\paragraph{Low-resource languages.} Apart from the four existing acceptability datasts, we also collected samples in 6 new languages, all annotated by theoretical syntacticians in their respective languages.  
These sentences are taken from five books/textbooks in the Cambridge Syntax Guides series, namely \textit{The Syntax of German}~\citep{syntaxGerman}, \textit{The Syntax of French}~\citep{syntaxFrench}, \textit{The Syntax of Spanish}~\citep{syntaxSpanish},  \textit{The Syntax of Arabic}~\citep{syntax-arabic} and \textit{The Syntax of Icelandic}~\citep{syntaxIcelandic}. Japanese data were collected from \textit{Handbook of Japanese Syntax}~\citep{syntaxJapanese}. 

Each book contains roughly one to three thousand example sentences with acceptability judgments made by linguists in respective languages. Graduate students majoring in linguistics in these languages were paid to extract all example sentences with their judgments in these books manually.   
Note that, following previous CoLA-style corpora, we only keep sentences labeled with ${}^{\texttt{*}}$ or ${}^{\texttt{??}}$ as our unacceptable sentences. All unmarked sentences are extracted as acceptable sentences.

Following previous acceptability datasets, we remove examples when the judgment is based on co-indexing of pronouns, empty categories, prosody or semantic/pragmatic interpretation. We also complete the sentence if it is composed of only a phrase, while keeping the judgment. 

For Japanese, we remove examples from its dialects (N=99) and those about classical Japanese (N=13).
For Arabic and Russian, as the original sentences are written in transliterations, we also convert them to their respective scripts manually. 

The mean time for data collection for one language is about a month, with Icelandic taking about 3 months as there were more examples in the book.

As these books/textbooks and handbooks are overviews of the syntax of each language, we believe they cover a wide range of linguistic phenomena in these languages, and can therefore serve as a good resource to evaluate language models' \textit{overall} ability to distinguish acceptable sentences from unacceptable ones.  

\subsection{Resulting Corpus and Data Split}

The resulting corpus contains more than 46k example sentences in 10 languages. 

For Italian and Chinese, we use the original train/dev/test splits of ItaCoLA and CoLAC, and for CoLAC we use the crowd label following \citet{2023CoLAC} 
For English and Russian, we keep the training splits of CoLA v.1.1 and RuCoLA, and use their in-domain development sets as our validation sets, and their out-of-domain development sets as our test sets.

For the six low-resource languages, 
we decide to adopt two splits for two purposes: fine-tuning smaller models such as XLM-R (\textit{v}1.0) and benchmarking LLMs (\textit{v}1.1).\footnote{Performance of LLMs on two splits of these data are similar (see Table~\ref{tab:split-compare}).}
For \textit{v}1.0, with the purpose for fine-tuning, we randomly sample 500 sentences from each of these languages to construct a training set, and divide the remaining sentences roughly equally between validation and test sets.
For \textit{v}1.1, we reserve 100 samples from each language as the validation set, and keep all the rest of the examples in the test set, thus producing a larger test set which we believe will make the evaluation more stable.
See Table~\ref{tab:mela} for details of the two splits.

\subsection{Evaluation Metric}
Following previous works in linguistic acceptability, we evaluate the performance on MELA by Matthews Correlation Coefficient~(MCC, \citealp{1975MCC}), which is a measure of similarity between binary distributions taking values from -1 to 1 and always yielding 0 for any two uncorrelated distributions, regardless of class imbalance.

\subsection{Comparison with Other Multilingual Benchmarks}

We note that all samples in MELA are constructed individually in each language. While some early multilingual benchmarks opt to translate English sentences into other languages to obtain parallel samples~\citep{2018XNLI,2019MLQA}, this approach does not suit our case.
First, the task of linguistic acceptability is highly language-dependent, and syntactic structures acceptable in one language may not be acceptable in another, and thus there is no easy way of translating existing corpora into other languages while keeping the target syntactic phenomena.
Second, as \citet{2020TyDiQA} and 
\citet{hu-kuebler-2021} argue, translation introduces artifacts into multilingual benchmarks and often results in translationese.

\begin{table*}[ht]
    \centering
    \adjustbox{width=\textwidth+0.5cm,center}{
    \begin{tabular}{lrrrrrrrrrrrrr}
    \toprule
        model & size & examples & en & zh & it & ru & de & fr & es & ja & ar & is & avg\\
    \midrule
        \rowcolor{gray!25}
        \multicolumn{14}{c}{\textbf{Supervised}} \\
        XLM-R & 550M & - & 60.64 & 54.94 & 53.53 & 49.37 & 26.72 & 19.04 & 34.08 & 29.32 & 14.12 & 35.41 & 37.72 \\
    \midrule
        \rowcolor{gray!25}
        \multicolumn{14}{c}{\textbf{Open-sourced}} \\
BLOOMZ$^0$ & 7.1B & - & 4.49 & 14.13 & 4.83 & 4.77 & 1.63 & 7.08 & 10.12 & 3.27 & 8.12 & 0.00 & 5.85 \\
BLOOMZ$^2$ & 7.1B & in-lang. & -1.11 & 7.65 & 5.67 & 5.38 & 3.90 & 5.19 & 6.76 & 3.83 & 6.22 & -0.35 & 4.31 \\
BLOOMZ$^2$ & 7.1B & en & -1.11 & 7.90 & 4.22 & 4.74 & 0.96 & 4.72 & 8.07 & 2.45 & 4.65 & -0.09 & 3.65 \\\midrule
mT0$^0$ & 13B & - & -5.42 & 9.68 & 12.68 & 5.57 & 11.33 & 8.24 & 2.88 & 13.20 & 6.77 & 1.22 & 6.62 \\
mT0$^2$ & 13B & in-lang. & 3.54 & 8.76 & 10.93 & 9.04 & 5.35 & 6.66 & 4.72 & 12.41 & 8.95 & 6.61 & 7.70 \\
mT0$^2$ & 13B & en & 3.54 & 8.06 & 10.22 & 10.68 & 7.17 & 7.40 & 6.09 & 10.46 & 2.91 & 4.86 & 7.14 \\\midrule
mTk$^0$ & 13B & - & 5.25 & -1.49 & -3.62 & 5.90 & 1.25 & 3.81 & 5.82 & 1.04 & 1.85 & 2.59 & 2.24 \\
mTk$^2$ & 13B & in-lang. & 22.74 & 8.47 & 10.24 & 16.66 & 11.96 & 9.28 & 13.34 & 12.00 & 4.87 & 10.93 & 12.05 \\
mTk$^2$ & 13B & en & 22.74 & 8.36 & 8.98 & 15.69 & 14.54 & 12.30 & 9.28 & 10.92 & 6.52 & 5.99 & 11.53 \\\midrule
Baichuan2-Base$^0$ & 13B & - & 0.00 & 0.00 & 0.00 & 0.00 & 0.00 & 0.00 & 0.00 & 0.00 & 0.00 & 0.00 & 0.00 \\
Baichuan2-Base$^2$ & 13B & in-lang. & 46.11 & 47.36 & 24.01 & 28.84 & 13.40 & 17.41 & 21.95 & 20.68 & 13.90 & 1.81 & 23.55 \\
Baichuan2-Base$^2$ & 13B & en & 46.11 & 35.16 & 17.84 & 25.88 & 6.42 & 15.95 & 16.57 & 13.48 & 11.41 & -3.80 & 18.50 \\\midrule
Baichuan2-Chat$^0$ & 13B & - & 37.15 & 33.56 & 10.08 & 7.93 & -6.49 & 8.41 & 18.32 & 11.15 & 0.00 & 2.01 & 12.21 \\
Baichuan2-Chat$^2$ & 13B & in-lang. & 41.12 & 29.25 & 18.10 & 19.46 & 6.46 & 18.57 & 20.81 & 14.18 & 13.97 & -1.51 & 18.04 \\
Baichuan2-Chat$^2$ & 13B & en & 41.12 & 27.02 & 12.22 & 14.11 & 2.80 & 9.49 & 14.62 & 11.40 & 7.00 & -5.03 & 13.47 \\\midrule
        \rowcolor{gray!25}
        \multicolumn{14}{c}{\textbf{Close-sourced}} \\
        GPT-3.5$^0$ &- & - & 64.60 & 17.01 & 14.33 & 18.05 & 23.01 & 31.66 & 24.35 & 16.61 & 9.57 & 4.69 & 22.39 \\
        GPT-3.5$^2$ &- & in-lang. & 64.11 & 25.32 & 38.66 & 21.59 & 21.62 & 29.52 & 44.20 & 21.48 & 6.19 & 9.70 & 28.24 \\
        GPT-3.5$^2$ &- & en & 64.11 & 30.25 & 25.27 & 24.91 & 24.54 & 29.88 & 37.75 & 21.70 & 6.43 & 0.56 & 26.54 \\
        \midrule
        GPT-4o$^0$ & - & - & 69.05 & \textbf{62.38} & 53.01 & 55.24 & 36.61 & 37.01 & 58.13 & 50.32 & 29.86 & 40.63 & 49.22 \\
        GPT-4o$^2$ &- & in-lang. & \textbf{72.14} & 59.01 & \textbf{54.86} & \textbf{59.17} & \textbf{39.66} & 37.19 & \textbf{61.36} & \textbf{52.03} & \textbf{32.38} & \textbf{43.64} & \textbf{51.14} \\
        GPT-4o$^2$ &- & en & \textbf{72.14} & 54.77 & 52.51 & 52.96 & 39.20 & \textbf{39.50} & 52.61 & 47.98 & 30.08 & 31.61 & 47.34 \\
    \bottomrule
    \end{tabular}
    }
    \caption{Performance of large language models on MELA, in comparison with XLM-R finetuned on MELA training set (all 10 languages). Superscripts denote the number of in-context examples. 
    Note that XLM-R is fine-tuned and evaluated on \textit{v}1.0 while LLMs are evaluated on \textit{v}1.1. However, the performance of LLMs on the two versions is consistent (see Table~\ref{tab:split-compare}). Thus we report results from different data splits in the same table. We also evaluate mTk with its origin CoLA prompt in its training set (see Table~\ref{tab:compare-mtk}). }
    \label{tab:llm-new}
\end{table*}
\section{Evaluating LLMs with MELA}\label{sec:llm}
In this section, we
report the performance of fine-tuned XLM-R and several LLMs, open-source or close-source, on MELA. 

\subsection{Experimental Settings}

To establish a supervised baseline, we use XLM-RoBERTa~\citep{2019XLM-R}, which is a multilingual version of RoBERTa~\citep{2019RoBERTa} pre-trained on 2.5TB CommonCrawl corpus covering one hundred languages.
XLM-R is fine-tuned on the combined training sets of all languages in MELA \textit{v}1.0 with the hyper-parameters described in Appendix~\ref{sec:appendix-training}.

For open-source LLMs, we consider BLOOMZ~\citep{2022BLOOM,2022xP3}, two instruction finetuned variants of mT5~\citep{2020mT5}---namely mTk~\citep{2022SuperNatural} and mT0~\citep{2022xP3}---and Baichuan2-Chat~\citep{2023Baichuan2} along with its base model. BLOOMZ is both pre-trained and fine-tuned on 46 languages, which only covers 5 languages in MELA: English, Chinese, French, Spanish, and Arabic\footnote{\citet{2022xP3} examine BLOOM's pre-training corpus ROOTS and estimate it to also contain a small amount of Russian, German, Italian, and Japanese.}. The pre-training corpus of mT5 includes all 10 languages in MELA, but mT0 is fine-tuned on the same instruction dataset as BLOOMZ. mTk's fine-tuning data, on the other hand, covers nine languages in MELA (except for Icelandic) and includes the English CoLA dataset. For Baichuan2, the exact language distribution of pre-training and fine-tuning data is not disclosed. For close-source models, we consider GPT-3.5 and GPT-4o~\citep{2023GPT4}.

There are several decisions to make when evaluating the above LLMs on MELA: prompt selection and the number of examples in the few-shot scenario. 
After some pilot experiments, which we describe in Appendix~\ref{sec:appendix-benchmark-details}, we opt to use a binary-choice method with the best performing prompt on the development set, and report the results on the test set in both zero-shot and two-shot scenarios.

\begin{table*}[!ht]
    \centering
    \adjustbox{width=\textwidth,center}{
    \begin{tabular}{ccccccccccc|c}
     \toprule
        $\downarrow$train (size) / eval$\rightarrow$ & en & zh & it & ru & de & fr & es & ja & ar & is & avg \\
    \midrule
        en (8551) & \textbf{71.66} & 47.41 & 28.23 & 31.91 & 24.85 & 18.96 & \textbf{32.21} & \textbf{34.50} & 21.50 & 24.47 & \textbf{33.57} \\
        zh (6072) & 45.72 & \textbf{52.71} & 23.18 & 22.80 & 21.31 & 17.61 & 29.01 & 31.48 & 22.16 & 20.57 & 28.65 \\
        it (7801)& 39.13 & 34.86 & \textbf{53.75} & 17.02 & 17.23 & 21.23 & 22.46 & 20.10 & 19.87 & 17.92 & 26.36\\
        ru (7869) & 50.29 & 39.77 & 24.26 & \textbf{47.22} & 20.47 & 14.11 & 28.62 & 32.48 & 20.11 & 24.49 & 30.18 \\
    \midrule
        de (500) & 35.87 & 37.97 & 15.44 & 18.38 & \textbf{36.13} & 16.45 & 22.06 & 22.68 & 12.27 & 21.67 & 23.89 \\
        fr (500)& 18.57 & 21.16 & 6.52 & 9.19 & 9.85 & \textbf{29.73} & 14.28 & 13.32 & 11.63 & 12.74 & 14.70 \\
        es (500)& 35.48 & 38.76 & 17.71 & 16.01 & 11.43 & 11.38 & 26.75 & 24.48 & 19.14 & 13.46 & 21.46 \\
        ja (500)& 22.67 & 20.32 & 10.20 & 12.40 & 13.82 & 10.44 & 10.81 & 33.62 & 8.85 & 11.21 & 15.43 \\
        ar (500)& 9.26 & 13.34 & 6.52 & 3.12 & 11.95 & 10.44 & 8.82 & 5.90 & \textbf{37.42} & 7.61 & 11.44 \\
        is (500)& 27.40 & 23.16 & 9.82 & 11.60 & 7.58 & 18.72 & 18.45 & 12.46 & 7.50 & \textbf{25.12} & 16.18 \\
    \midrule
        avg. high-resource & 51.70 & 43.69 & 32.35 & 29.74 & 20.96 & 17.98 & 28.07 & 29.64 & 20.91 & 21.86 & 29.69\\
        avg. low-resource & 24.88 & 25.79 & 11.04 & 11.78 & 15.13 & 16.19 & 16.86 & 18.74 & 16.14 & 15.30 & 17.18\\
        avg. w.o. in-lang. & 31.60 & 30.75 & 15.76 & 15.83 & 15.39 & 15.48 & 20.75 & 21.93 & 15.89 & 17.13 & -\\
    \bottomrule
    \end{tabular}
    }
    \caption{Cross-lingual transfer results of finetuned XLM-R. The top four training languages are high-resource languages in MELA (whose training samples vary from 6000 to 8500). The middle six are low-resource languages in MELA (all of which have 500 training samples). All results are the median MCC of seven runs. ``Avg. high-resource'' refers to the average of the first four rows, while ``avg. low-resource'' is the average of the next six rows. To illustrate the effects of in-language training, figures in the last row are the average MCC on each language's validation set of 9 rows, except the one where the model is trained in-language.}
    \label{tab:transfer}
\end{table*}
\subsection{Main results}
Results of fine-tuned XLM-R and LLMs evaluated on MELA are given in Table~\ref{tab:llm-new}. We make the following observations.

\paragraph{Observation 1: GPT-4o exhibits a strong multilingual ability for acceptability judgement.}
It achieves the best performance on each individual language in MELA, exceeding supervised fine-tuned XLM-R. 
Its performance is 11 points higher than finetuned XLM-R even in a 0-shot setting. 
There is a bigger gap between GPT-4o and finetuned XLM-R on low-resource languages than high-resource ones, likely due to the small amount of training data (500 examples) for XLM-R. 
Compared to GPT-4o, GPT-3.5 seems to be more English-centric, with drastic performance drop in non-English languages. 

\paragraph{Observation 2: LLMs benefit more from in-language examples in two-shot setting.}
Our results suggest that prompting with two English examples (most of the time) leads to a lower performance than prompting with in-language examples. 
On Icelandic, for example, the MCC of the 2-shot setting with English in-context examples is even lower than the 0-shot performance for Baichuan2, GPT-3.5, and GPT-4o. 

\paragraph{Observation 3: Baichuan2-Base requires in-context examples}
As shown in Table~\ref{tab:llm-new}, under zero-shot setting, Baichuan2-Base shows random performance\footnote{Baichuan2-Base always chooses ``B. Unacceptable'' as the answer, under all prompts we tested.}, while its Chat model exhibits non-trivial performance, even for languages such as Spanish. The Base model benefits more from in-context learning examples though, surpassing the Chat model in two-shot settings.

\begin{figure*}[th]
    \centering
    \includegraphics[width=1\textwidth]{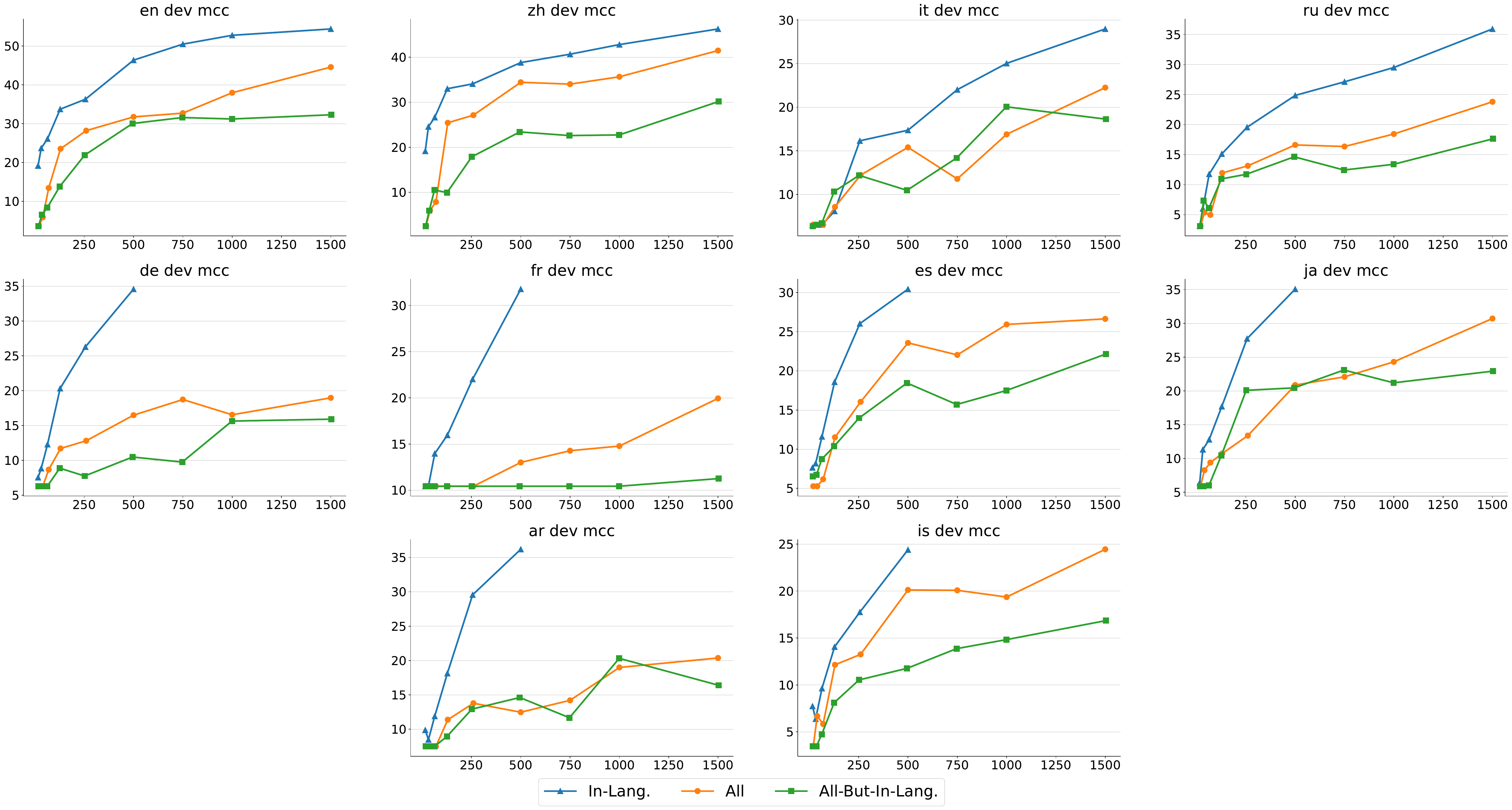}
    \caption{Performance of XLM-R when fine-tuned on different languages. The horizontal axis indicates the number of training samples. For example, for ``all'' curves, the point at 500 indicates the model is trained on 500 sentences, with 50 from each language. For ``All-but-in-lang.'' curves, the point at 495 indicates the model is trained on 495 sentences, with 55 from each of the nine languages except the one being evaluated on.}
    \label{fig:mft}
\end{figure*}

\section{Cross-lingual Transfer and Multilingual Fine-tuning}\label{sec:transfer}
In this section, we investigate cross-lingual transfer and multilingual fine-tuning of linguistic acceptability with XLM-R. 
All training and evaluation are done on MELA \textit{v}1.0 as it requires a training set. 

\subsection{Experimental Settings}

\paragraph{Cross-lingual Transfer} 
To observe the transfer of acceptability judgements across languages, we fine-tune XLM-R on one language, and evaluate on all 10 development sets. We report the median MCC of seven runs for all results to mitigate inter-run variance.\footnote{Training details can be found in Appendix~\ref{sec:appendix-training}. }

\paragraph{Multilingual Fine-tuning}
We downsample sentences in each language to the same number, and fine-tune XLM-R in three settings: (1) in-language, where the fine-tuning and evaluation languages are the same; (2) all-language, where the model is fine-tuned on a mixture of data containing an equal number of sentences from ten languages; and (3) all-but-in-language, where the model is fine-tuned on a mixture of data containing an equal number of sentences from nine languages, except the one being evaluated on.

\subsection{Results}

Results for cross-lingual transfer and multilingual finetuning of XLM-R are reported in Table~\ref{tab:transfer} and  Figure~\ref{fig:mft}.
We make the following  observations.

\paragraph{Observation 1: Cross-lingual transfer is non-trivial. } In Table~\ref{tab:transfer}, we see that all numbers are (much) greater than 0, suggesting that transferring from language A to language B is possible, even for acceptability judgment tasks. For instance, fine-tuning XLM-R on 500 Icelandic examples results in 23.16 MCC for a completely unrelated language, Mandarin Chinese. 
A similar conclusion can be drawn from Figure~\ref{fig:mft}, where the green line, which has no in-language training data, demonstrates an increasing trend for all languages, except for French, which plateaus at around 10 MCC. 

\paragraph{Observation 2: Size of training set matters, but not always. }
The overall performance when high-resource languages are used as training data ($>$6k training examples), 29.69 MCC, is higher than when low-resource ones are used ($=$500 examples), 18.18 MCC, as shown in the last block of Table~\ref{tab:transfer}.
However, it must be pointed out that sometimes a (14 times) larger training set does not lead to better performance. For instance, in the second column of Table~\ref{tab:transfer}, when evaluated on Chinese, 500 examples of German or Spanish achieve roughly 37 MCC, which is on par with having more than 7,000 training examples for Italian and Russian, with 34.86 and 39.77 MCC respectively. 
Similarly, when evaluated on Icelandic, 500 German examples again demonstrate a performance on par with many thousands of Chinese, Italian or Russian examples (second to last column of Table~\ref{tab:transfer}).
Thus the transferring performance between two languages seems to be a result of both the language pair in question as well as the size of the training set. 

\paragraph{Observation 3: Among low-resource languages, Arabic training data has the lowest average performance (Table~\ref{tab:transfer}).} This is likely due to the fact that Arabic is from a different language family from all other nine languages. 
From the last column of Table~\ref{tab:transfer}, we observe that German and Spanish training data have the best performance, likely because MELA has three Germanic languages and three Romance languages, which may make cross-lingual transfer easier among these cognate languages. 

\paragraph{Observation 4: For Italian, Japanese and Arabic, during multi-task training, adding in-language data does not affect performance much. } 
From Figure~\ref{fig:mft}, we see that the green and orange lines cross for these three languages, suggesting that \textit{when training with mixed-language data},  in-language examples may not be very critical for these languages.

\begin{table}[ht]
\vspace{-0.3cm}
\centering
\adjustbox{width=\linewidth}{
\begin{tabular}{cccccc}
\\\toprule
Task & base & en & it & ru & zh \\\midrule
\texttt{pos} & 92.87 & $+$0.90 & $+$0.60 & $+$0.30 & $+$1.08 \\
\texttt{dep} & 89.41 & $+$0.93 & $+$0.72 & $+$0.51 & $+$0.45 \\
\texttt{const} & 78.54 & $+$0.56 & $-$0.10 & $+$0.72 & $+$0.42 \\
\texttt{name} & 93.49 & $+$0.74 & $-$0.15 & $+$1.04 & $+$0.59 \\
\texttt{srl} & 77.93 & $+$4.41 & $+$2.07 & $+$3.31 & $+$2.35 \\
\texttt{coref} & 83.84 & $+$1.71 & $+$0.28 & $+$0.14 & $+$0.69 \\\midrule
\texttt{avg} & 86.01 & $+$1.54 & $+$0.57 & $+$1.00 & $+$0.93\\\bottomrule
\end{tabular}}
\caption{We report the F1 score of XLM-R$_{base}$ (base) on each probing task and the differences between the probing results between MELA-fine-tuned XLM-Rs (en, it, ru and zh) and the base model.}
\label{tab:edge-probing-tasks}
\end{table}

\begin{table*}[ht]
\centering
\adjustbox{width=.95\textwidth,center}{
\begin{tabular}{cl||ccccc|cccccc}
\toprule
\multicolumn{2}{c||}{Probing task} & \multicolumn{5}{c|}{Part-of-speech tagging} & \multicolumn{5}{c}{Depedency labeling} \\\midrule
\multicolumn{2}{c}{$\downarrow$eval / train$\rightarrow$} & en & it & ru & zh &\textbf{avg} & en & it & ru & zh &\textbf{avg} \\\midrule
\multirow{2}{*}{en} & M$_{base}$ & 92.87& 75.77& 65.63& 43.33 &69.40& 89.41& 74.99& 60.67& 40.05&66.28\\
  & M$_{mela}^{en}$ & 93.77& 81.43& 68.22& 44.66 &\textbf{72.02}& 90.34& 77.40& 61.84& 45.44&\textbf{68.76}\\\hline
\multirow{2}{*}{it} & M$_{base}$ & 83.26& 94.61& 66.90& 38.73 &70.88& 78.17& 91.50& 60.65& 32.35&65.67\\
 & M$_{mela}^{it}$ & 85.60& 95.71& 63.73& 39.70 &\textbf{71.19}& 83.56& 92.46& 62.85& 37.31&\textbf{69.05}\\\hline
\multirow{2}{*}{ru} & M$_{base}$ & 82.97& 79.90& 95.53& 53.18 &77.90& 77.72& 78.86& 90.90& 42.77&72.56\\
 & M$_{mela}^{ru}$ & 85.42& 81.01& 95.43& 54.06 &\textbf{78.98}& 80.65& 81.27& 92.04& 46.10&\textbf{75.02}\\\hline
\multirow{2}{*}{zh} & M$_{base}$ & 61.19& 58.57& 64.43& 93.88 &69.52& 50.16& 43.42& 43.12& 86.06&55.69\\
 & M$_{mela}^{zh}$ & 64.55& 55.60& 63.98& 94.35 &\textbf{69.62}& 55.42& 44.52& 44.16& 87.73&\textbf{57.96}\\\bottomrule
\end{tabular}}
\caption{F1 scores of Experiment 2 on part-of-speech tagging and depedency labeling in a cross-lingual setting. M$_{base}$ refers to the pre-trained XLM-R model; M$^{en}_{mela}$ refers the XLM-R fine-tuned on the English MELA. 
\textbf{Bold} denotes a better performance in average between M$_{base}$ and M$_{mela}^{lang}$. 
We conduct a pair comparison between M$_{base}$ and M$_{mela}^{lang}$ trained on MELA of one language to investigate whether linguistic acceptability helps the cross-lingual transfer in the above two probing tasks.
For each cell, probing classifiers are trained on span representations in the language denoted in the second row, which are encoded by the model denoted in the second column, and evaluated on the probing tasks in the same language on which XLM-R is fine-tuned (the first column). 
}
\label{tab:cross-pos-dep}
\end{table*}

\section{Edge Probing}
\label{sec:analysis}

In this section, we adopt edge probing~\citep{2019TennyWeighted,2019TennyEdge} to explore whether fine-tuning on acceptability judgment tasks injects  syntax-related information into the pre-trained XLM-R.

\subsection{Experimental settings}

Edge probing focuses on structural labeling tasks in the form of span labeling. 
We choose following tasks: 1) part-of-speech tagging, 2) dependency labeling, 3) constituency labeling, 4) named entity labeling, 5) semantic role labeling, and 6) co-reference.\footnote{Tasks 1-2 are from UD~\citep{2021UD}; Tasks 3-6 are from OntoNotes~\citep{2013OntoNotes}.} 
Take dependency labeling as an example, representations of a dependent and its head, encoded by an XLM-RoBERTa, are used to train a probe classifier to predict the dependency relation between the two words.

We hypothesize that training on MELA can improve the performance of XLM-R on the syntax-related probing tasks above, and design the following experiments.

\paragraph{Experiment 1}
We train probing classifiers using span representations from XLM-Rs on English probing tasks.
We set the pre-trained M$_{base}$ (pre-trained XLM-R) as the control group, and the other four MELA-fine-tuned M$^{lang}_{mela}$ as the test group, where 
$lang$ specifies the training data of which language from MELA were used for fine-tuning.

\paragraph{Experiment 2}
For two tasks (\texttt{pos} and \texttt{dep}) with multilingual data available, we experiment on cross-lingual transfer as well. We train probing classifiers on representations from M$_{base}$ in each of four high-resource languages and run zero-shot evaluation on a target language (\textit{lang}). We repeat the procedure on M$^{lang}_{mela}$ (see more details in Appendix~\ref{append:edge-probing}).

\subsection{Results}

In Experiment 1 we train probing classifiers using representations from different XLM-R variants, some of which have been fine-tuned on MELA while the base model has not. Results in Table~\ref{tab:edge-probing-tasks} show that the average performance of XLM-R$_{base}$ on the six probing tasks is the lowest across the six edge probing tasks (see the last row). 
We further observe from Table~\ref{tab:edge-probing-tasks} that the semantic role labeling task benefits most from MELA-fine-tuning.

In Experiment 2, the performances of probing tasks are evaluated in the cross-lingual transfer setting. 
We compare the the pre-trained XLM-R model (M$_{base}$) and XLM-Rs fine-tuned on the linguistic acceptability judgement task of a specific language (M$_{mela}^{lang}$) (see Table~\ref{tab:cross-pos-dep}). 
The results indicate that the probing classifiers trained on span representations from fine-tuned XLM-R models achieve better performance than the base model. 
Fine-tuning on one language of MELA helps the model transfer to that language, and more often than not other languages in part-of-speech tagging and dependency labeling. 

MELA-fine-tuned XLM-Rs perform better on syntax-related probing tasks in mono-lingual and cross-lingual settings, supporting our hypothesis.

\section{Conclusion}\label{sec:conclusion}

In this work we present MELA, the first multilingual acceptability judgement benchmark covering a diverse set of 10 languages, all annotated by expert linguists. By benchmarking multilingual LLMs on MELA and fine-tuning XLM-R in different cross-lingual settings, we find that (1) GPT-4o ourperforms supervised XLM-R, especially on low-resource languages, that (2) in-language data is crucial for few-shot evaluation and that (3) cross-lingual transfer is non-trivial for all language pairs in supervised fine-tuning. We probe MELA-fine-tuned XLM-R for the syntax information encoded, finding that training on MELA improves the performance on syntax-related probing tasks, which indicates that language models acquire syntactic knowledge during training on linguistic acceptability judgements.

\section*{Limitations}
Due to the large amount of human labor involved in transcribing and examining the sentences in MELA, the dataset only covers ten languages, of which six are low-resource, with only a small number of training samples. In the future, we intend to expand the dataset by additionally collecting data in other languages, especially non-Latin and non-Indo-European languages, which are currently underrepresented in MELA.

Also, in this work we focused on introducing the MELA dataset and showcasing some of its usages, such as benchmarking LLMs and providing a data resource for cross-lingual research in computational linguistics. 
We leave the exploration of other use cases of MELA to future work.

\section*{Ethics Statement}
Sentences in our dataset MELA, including those in English, Italian, Russian, and Chinese consolidated from previous works, are sourced from renounced linguistics publications such as syntax textbooks and journal articles. Therefore, we believe they do not raise any ethical issues such as leak of personal identifiable information.

The sentences in MELA, both acceptable and unacceptable, are only intended for research concerning the acquisition and evaluation of linguistic capabilities (of either humans or language models), and should not be interpreted otherwise. 
For individual sentences in MELA, the copyright (where applicable) remains with the original authors or publishers.
We ask researchers who use MELA to also cite the original source, i.e., CoLA~\citep{2018CoLA}, ItaCoLA~\citep{2021ItaCoLA}, RuCoLA~\citep{2022RuCoLA} and CoLAC~\citep{2023CoLAC}.

\section*{Acknowledgments}

We thank Bingjie Mao, Jinchi Jiang, Wen Shi, Jialin Guo, Chenhui Liu and Licen Liu for their help in data collection. 
We also appreciate the suggestions and comments from the anonymous reviewers. 
This study is supported by Shanghai Pujiang Program awarded to Hai Hu (22PJC063).
Ziyin Zhang and Rui Wang are partially supported by the National Natural Science Foundation of China (62176153) and the Shanghai Municipal Science and Technology Major Project (2021SHZDZX0102, as the MoE Key Lab of Artificial Intelligence, AI Institute, Shanghai Jiao Tong University).

\bibliography{custom}

\appendix

\section{Benchmark Details} \label{sec:appendix-benchmark-details}
In this section, we provide details about how we make decisions to benchmark LLMs, which includes prompt selection, and the number of examples in the few-shot scenario, along with details of fine-tuning XLM-R.

\subsection{Large Language Models}
\label{sec:appendix-llm}

\begin{figure*}[ht]
    \centering
    \includegraphics[width=\linewidth]{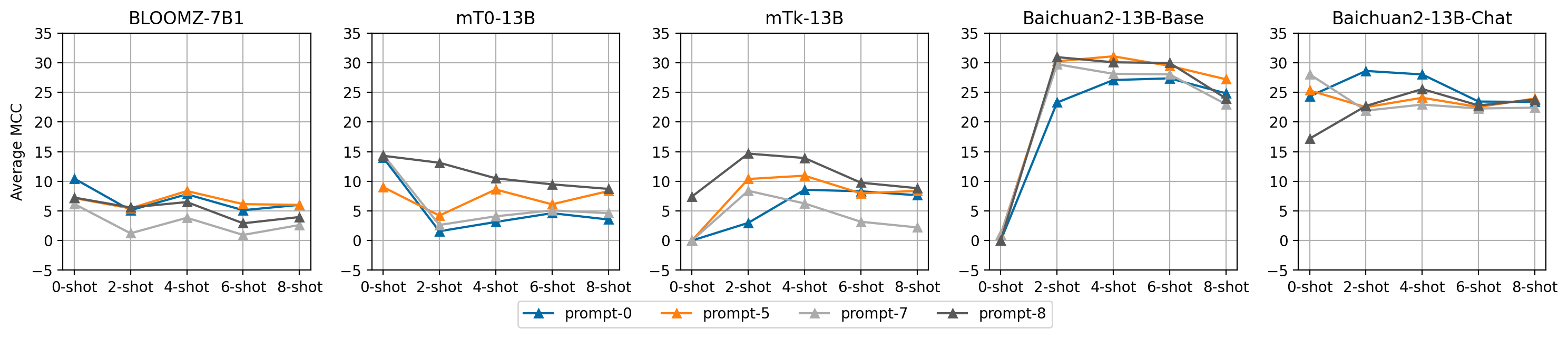}
    \caption{Prompt selection results. We experiment with 4 prompts adapted from previous CoLA-prompts from \href{https://github.com/bigscience-workshop/promptsource/blob/main/promptsource/templates/glue/cola/templates.yaml}{\texttt{promptsource}} and \href{https://github.com/EleutherAI/lm-evaluation-harness/blob/main/lm_eval/tasks/glue/cola/default.yaml}{\texttt{lm-evaluation-harness}}.}
    \label{fig:mulchoice:4prompt}
\end{figure*}

\begin{figure}[ht]
    \centering    \includegraphics[width=\linewidth]{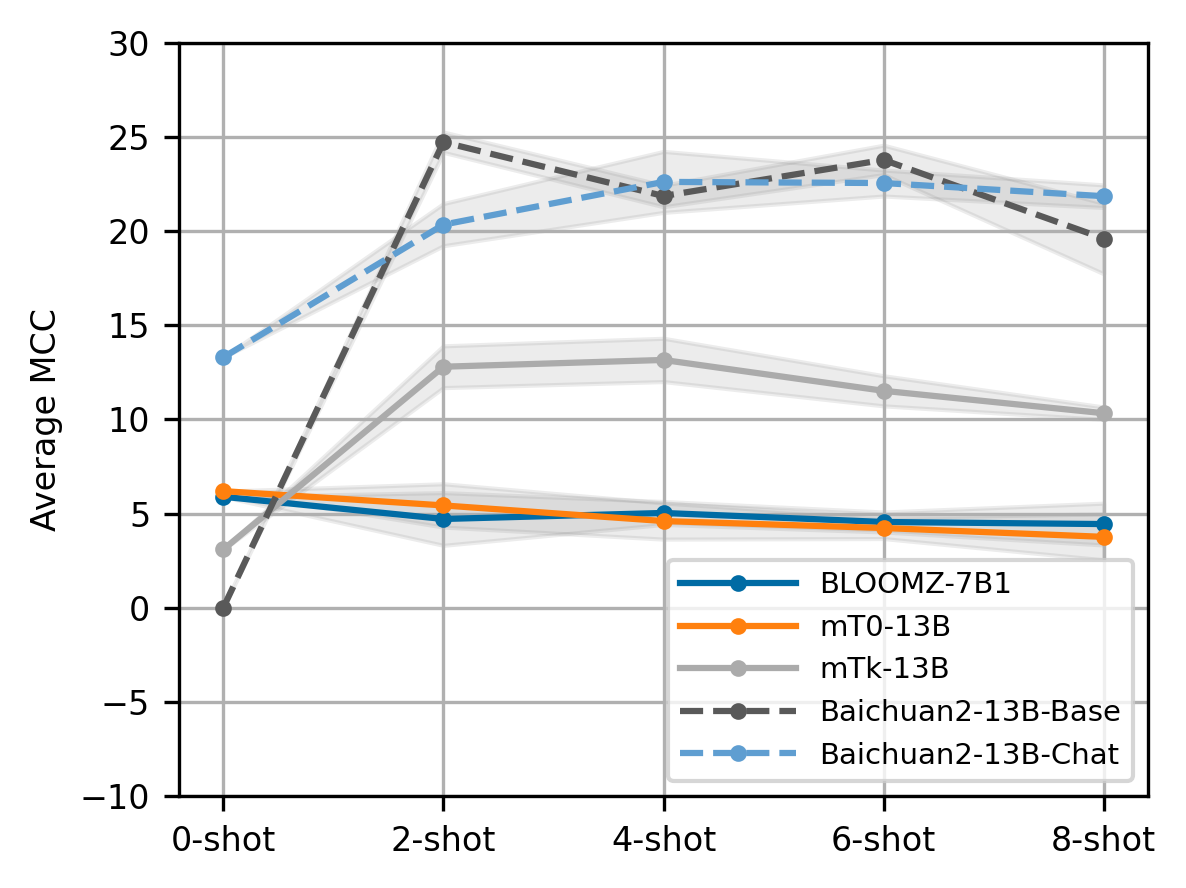}
    \caption{Average performance across languages with different numbers of in-context examples. We average the MCC and report standard deviations over 5 seeds. Gray bands denote standard deviations.}
    \label{fig:mulchoice:1prompt}
\end{figure}

Following MMLU~\cite{2020MMLU}, we evaluate MELA in the multiple-choice format. 
We use prompts that end with ``Answer:''. 
Models are required to produce probabilities for index tokens ``A'' and ``B''. 
The index token with a higher probability is regarded as the decision of models. 
We tune prompts with open-sourced LLMs on the validation set in our pilot experiment. 

We first experiment on 1,000 samples of MELA (50 samples per label per language), using 4 different prompts in 0/2/4/6/8-shot (equal number of positive and negative examples) scenarios with in-language examples provided. 
The prompt with the highest average MCC is selected to evaluate LLMs on the whole MELA test set (see Figure~\ref{fig:mulchoice:4prompt} and~\ref{fig:mulchoice:1prompt}). 
The results indicate that 1) \texttt{prompt-8} is better than others and 2) models no longer improve with more than 2 in-context examples. 

Therefore, we carry out formal experiments with \texttt{prompt-8} (see Figure~\ref{fig:prompts-mtk}) for both open-sourced LLMs and GPT models in zero and two-shot scenarios (see~\S\ref{sec:llm}). 

\begin{table*}[ht]
    \centering
    \adjustbox{width=\textwidth+0.5cm,center}{
    \begin{tabular}{clrrrrrrrrrrrrr}
    \toprule
    prompt & model & size & examples & en & zh & it & ru & de & fr & es & ja & ar & is & avg\\
    \midrule
\multirow{2}{*}{\texttt{ours}} & mTk$^2$ & 13B & in-lang. & 22.74 & 8.47 & 10.24 & 16.66 & 11.96 & 9.28 & 13.34 & 12.00 & 4.87 & 10.93 & 12.05 \\
& mTk$^2$ & 13B & en & 22.74 & 8.36 & 8.98 & 15.69 & 14.54 & 12.30 & 9.28 & 10.92 & 6.52 & 5.99 & 11.53 \\\midrule
    \multirow{2}{*}{\texttt{origin}} & mTk$^2$ & 13B & in-lang. & 39.13 & 32.18 & 18.26 & 11.83 & 9.91 & 13.09 & 24.42 & 22.45 & 12.72 & 15.54 & 19.95\\
    & mTk$^2$ & 13B & en & 39.13 & 31.48 & 12.12 & 14.92 & 16.46 & 12.81 & 15.77 & 15.17 & 6.34 & 11.21 & 17.54 \\
    \bottomrule
    \end{tabular}
    }
    \caption{We compare the results of mTk on our prompt and the origin prompt in its training data.}
    \label{tab:compare-mtk}
\end{table*}

Note that mTk includes the 2-shot CoLA task in its training data. 
We also reuse the prompt for Supernatural Instruction task 616\footnote{\url{https://github.com/allenai/natural-instructions/blob/master/tasks/task616_cola_classification.json}}. 
We compare the results of mTk's origin CoLA prompt (see Figure~\ref{fig:prompts-mtk}) and our multiple-choice prompt (see Table~\ref{tab:compare-mtk}).

\lstset{
  basicstyle=\normalsize,
  columns=fullflexible,
  frame=single,
  breaklines=true,
  breakindent=0pt,
}

\begin{figure*}
\centering
\begin{lstlisting}
# 0-shot multiple-choice prompt
Determine whether the following sentence(s) violate certain linguistic constraints. If yes, then it is "unacceptable"; otherwise, "acceptable".

Sentence: {target sentence}.
Determine whether this sentence is acceptable or unacceptable?
A. Acceptable
B. Unacceptable
Answer: 
\end{lstlisting}
\begin{lstlisting}
# 2-shot multiple-choice prompt
Determine whether the following sentence(s) violate certain linguistic constraints. If yes, then it is "unacceptable"; otherwise, "acceptable".

Sentence: {positive example1}.
Determine whether this sentence is acceptable or unacceptable?
A. Acceptable
B. Unacceptable
Answer: A

Sentence: {negative example2}.
Determine whether this sentence is acceptable or unacceptable?
A. Acceptable
B. Unacceptable
Answer: B

Sentence: {target sentence}.
Determine whether this sentence is acceptable or unacceptable?
A. Acceptable
B. Unacceptable
Answer: 
\end{lstlisting}
\begin{lstlisting}
# 2-shot mTk origin prompt
Definition: You're given a sentence and your task is to classify whether the sentence is acceptable or not. Any sentence which is grammatically correct, has a naturalistic text, is written by a native speaker and which minimizes superfluous content is acceptable, otherwise unacceptable. If the sentence is acceptable then write "acceptable", otherwise "unacceptable".
Positive Example 1-
	input: {positive example1}
	output: acceptable
Positive Example 2-
	input: {negative example2}
	output: unacceptable
Now complete the following example-
	input: {target sentence}
	output:
\end{lstlisting}
\caption{Prompt used for evaluating LLMs.}
\label{fig:prompts-mtk}
\end{figure*}

\subsection{XLM-R Fine-tuning Details}\label{sec:appendix-training}
For experiments concerning XLM-R in \S\ref{sec:llm} and \S\ref{sec:transfer}, we finetune with learning rate 7.5e-6, weight decay 0.075 and batch size 32. To minimize confounding variables and accentuate the interaction across languages in terms of linguistic acceptability performance, we train the model for 5k steps for all experiments in \S\ref{sec:llm} and \S\ref{sec:transfer} with 750 steps of linear warmup and cosine learning rate decay over 0.4 cycles, and take the best checkpoint based on validation results. 

We note that these hyperparameters are chosen based on previous works on similar tasks~\citep{2019RoBERTa,2023CoLAC} and our preliminary experiments. The sheer amount of experiments covered in our work makes it impossible to finetune hyperparameters on each combination of training data, and we thus decide to keep them fixed across all experiments for a fair comparison across languages, which may be suboptimal for certain cases. \citet{2023CoLAC}, for example, report  56.45 MCC for XLM-R on CoLAC development set, while our result is 52.71 with the same training data.

\begin{figure*}[ht]
    \centering
    \includegraphics[width=1\textwidth]{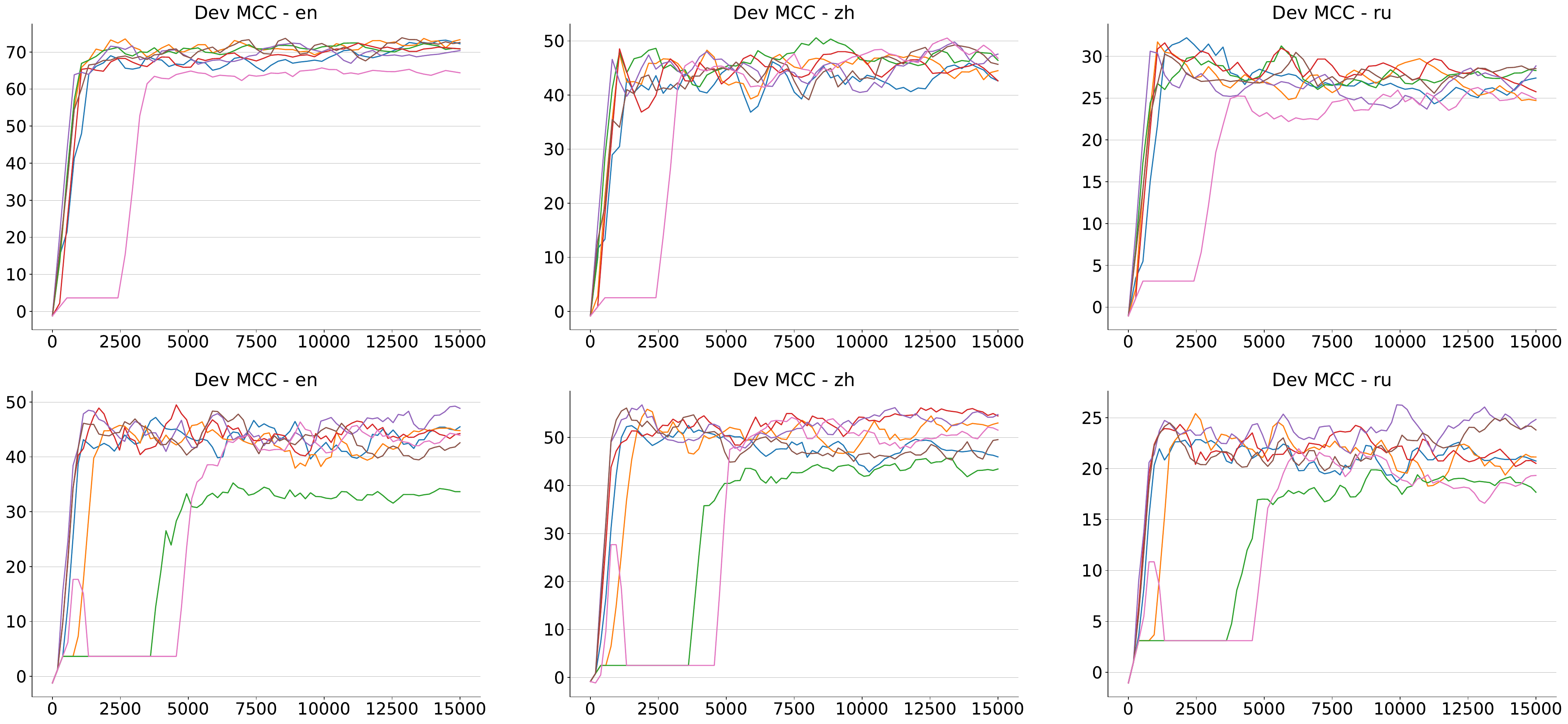}
    \caption{Interrun variance when finetuning XLM-R on English (first row) and Chinese (second row) training data. Each subfigure plots the validation MCC of seven runs with different random seeds on one language. After taking the median of these seven runs, this variance is mitigated to a large extent.}
    \label{fig:mcc_variance}
\end{figure*}

We also note that finetuning language models on linguistic acceptability data leads to large performance variations, regardless of the specific languages (see Figure~\ref{fig:mcc_variance}), which corresponds with previous findings in the literature~\citep{2019T5}. We thus train with seven different random seeds for every experiment in this work to reduce this variance, and the reported scores are computed by first taking the median of these seven runs at each checkpointing step, and then maxing over all the aggregated checkpoints. For experiments on downsampled data in \S\ref{sec:transfer}, each run also selects a different subset of training data.

\begin{table*}[ht]
\centering
\adjustbox{width=\textwidth+1cm,center}{
    \begin{tabular}{lrrrrrrrrrrrrrrrr} \toprule
        \multicolumn{2}{l}{model} & \multicolumn{3}{c}{BLOOMZ} & \multicolumn{3}{c}{mT0} & \multicolumn{3}{c}{mTk} & 
         \multicolumn{3}{c}{Baichuan2-Base} & \multicolumn{3}{c}{Baichuan2-Chat} \\\cmidrule(lr){3-5} \cmidrule(lr){6-8} \cmidrule(lr){9-11} \cmidrule(lr){12-14} \cmidrule(lr){15-17}
        \multicolumn{2}{l}{\textit{n}-shot} & 0-shot & 2-shot & 2-shot & 0-shot & 2-shot & 2-shot & 0-shot & 2-shot & 2-shot & 0-shot & 2-shot & 2-shot & 0-shot & 2-shot & 2-shot \\
        \multicolumn{2}{l}{ex. lang.} & - & in-lang. & en & - & in-lang. & en & - & in-lang. & en & - & in-lang. & en & - & in-lang. & en \\\midrule
de$_{v1.0}$ & 273 & -12.22 & 0.46 & 1.71 & 7.00 & 8.36 & 6.83 & 16.63 & 10.70 & 10.70 & 0.00 & 7.98 & 2.77 & 0.00 & 0.86 & -3.05 \\
de$_{v1.1}$ & 945 & 1.63 & 3.90 & 0.96 & 11.33 & 5.35 & 7.17 & 1.25 & 11.96 & 14.54 & 0.00 & 13.40 & 6.42 & -6.49 & 6.46 & 2.80 \\\midrule
fr$_{v1.0}$ & 467 & 11.32 & 2.83 & 4.84 & 5.83 & 4.91 & 8.55 & 2.14 & 8.95 & 7.71 & 0.00 & 17.82 & 13.87 & 9.58 & 16.99 & 10.03 \\
fr$_{v1.1}$ & 1333 & 7.08 & 5.19 & 4.72 & 8.24 & 6.66 & 7.40 & 3.81 & 9.28 & 12.30 & 0.00 & 17.41 & 15.95 & 8.41 & 18.57 & 9.49 \\\midrule
es$_{v1.0}$ & 293 & 10.83 & 9.20 & 8.38 & 7.18 & 11.10 & 10.44 & 8.65 & 21.07 & 17.67 & 0.00 & 24.74 & 21.19 & 20.85 & 24.35 & 14.29 \\
es$_{v1.1}$ & 988 & 10.12 & 6.76 & 8.07 & 2.88 & 4.72 & 6.09 & 5.82 & 13.34 & 9.28 & 0.00 & 21.95 & 16.57 & 18.32 & 20.81 & 14.62 \\\midrule
ja$_{v1.0}$ & 581 & 1.98 & -0.31 & 2.19 & 12.32 & 12.58 & 9.51 & -2.80 & 5.58 & 5.70 & 0.00 & 24.93 & 16.72 & 9.06 & 20.92 & 14.96 \\
ja$_{v1.1}$ & 1561 & 3.27 & 3.83 & 2.45 & 13.20 & 12.41 & 10.46 & 1.04 & 12.00 & 10.92 & 0.00 & 20.68 & 13.48 & 11.15 & 14.18 & 11.40 \\\midrule
ar$_{v1.0}$ & 259 & 5.53 & 6.83 & 7.43 & 2.95 & 5.66 & -3.59 & 10.60 & 3.48 & 6.77 & 0.00 & 10.75 & 6.79 & 0.00 & 12.66 & 0.52 \\
ar$_{v1.1}$ & 917 & 8.12 & 6.22 & 4.65 & 6.77 & 8.95 & 2.91 & 1.85 & 4.87 & 6.52 & 0.00 & 13.90 & 11.41 & 0.00 & 13.97 & 7.00 \\\midrule
is$_{v1.0}$ & 899 & 0.00 & 0.91 & 0.95 & 3.85 & 4.08 & 4.18 & 7.74 & 14.04 & 6.81 & 0.00 & 3.99 & 0.75 & 2.68 & 3.14 & -3.08 \\
is$_{v1.1}$ & 2198 & 0.00 & -0.35 & -0.09 & 1.22 & 6.61 & 4.86 & 2.59 & 10.93 & 5.99 & 0.00 & 1.81 & -3.80 & 2.01 & -1.51 & -5.03 \\\midrule
avg$_{v1.0}$ & - & 2.91 & 3.32 & 4.25 & 6.52 & 7.78 & 5.99 & 7.16 & 10.64 & 9.23 & 0.00 & 15.03 & 10.35 & 7.03 & 13.15 & 5.61 \\
avg$_{v1.1}$ & - & 5.04 & 4.26 & 3.46 & 7.27 & 7.45 & 6.48 & 2.73 & 10.40 & 9.93 & 0.00 & 14.86 & 10.01 & 5.57 & 12.08 & 6.71 
\\\bottomrule
    \end{tabular}
    }
    \caption{Comparison between the performance of open-sourced LLMs on two versions of data splits. 
    We only report results on six low-resource languages since data for the four high-resource languages are the same between the two splits.}
    \label{tab:split-compare}
\end{table*}
\subsection{Comparison Between Two Splits} \label{sec:appendix-splits}
In \S\ref{sec:mela}, we have to split the MELA datasets into train, development, and test sets to fine-tune XLM-R. 
However, considering training data is no longer necessary for LLM evaluation. 
Therefore, we decide to make two different data splits. 
On the one hand, we want a fair comparison between supervised fine-tuning XLM-R and zero/few-shot LLMs. 
On the other hand, we want MELA to better fit the recent paradigm of LLM evaluation. 
In this case, we provide a comparison between the performance of LLMs on the two different versions of data (see Table~\ref{tab:split-compare}).
The results of the two versions are similar, by which we make it comparable between fine-tuned XLM-R on \texttt{v1.0}-test set and LLMs on \texttt{v1.1}-test set.

\begin{table*}[ht]
\centering
\begin{tabular}{lcccc}
\toprule
Task & $|L|$  & Sentences            & Words               & Total Labels          \\\midrule
Part-of-speech             & 17 & 0.7k / 0.15k / 0.15k & 14.7k / 3.2k / 3.3k & 14.7k / 3.2k / 3.3k   \\
Dependencies               & 36 & 0.7k / 0.15k / 0.15k & 14.7k / 3.2k / 3.3k & 14.7k / 3.2k / 3.3k   \\
Constituencies             & 78 & 1.4k / 0.3k / 0.3k   & 27.0k / 5.9k / 5.7k & 51.1k / 11.1k / 10.7k \\
Named Entities             & 18 & 1.4k / 0.3k / 0.3k   & 34.6k / 7.3k / 7.4k & 3.7k / 0.8k / 0.7k    \\
Semantic Roles             & 2  & 1.4k / 0.3k / 0.3k   & 29.9k / 6.4k / 6.6k & 7.3k / 1.5k / 1.6k    \\
Co-reference               & 66 & 1.4k / 0.3k / 0.3k   & 35.4k / 8.1k / 7.5k & 3.6k / 0.8k / 0.7k   \\\bottomrule
\end{tabular}
\caption{The summary statistics for each split and for each English probing task.}
\label{tab:probe-dataset}
\end{table*}

\section{Edge Probing Details}
\label{append:edge-probing}
\paragraph{Probing Classifier} We follow the same architecture of probing classifier as~\citep{2019TennyEdge}.
We extract contextual representations from each layer of XLM-R (including the embedding layer), and get the scalar mixed representations (in 1,024-dim), see Equation (1) in~\citep{2019TennyWeighted}. Then, the representations are projected in 512-dim with a CNN module. 
For two-span prediction, we concatenate representations of two spans into a 1,024-dim tensor.
We pass the span representations to the probing classifier, which is a two-layer MLP (hidden state dimension is set to 512). 

\paragraph{Probing Dataset}
For part-of-speech tagging and dependency labeling, we use PUD (parallel sentences in all four languages) in UD V2.13. For the other four tasks in OntoNotes 5.0, we downsample sentences to 2k. All datasets are split into train, development and test sets in a ratio of 7:1.5:1.5. For each sentence, there might be multiple labels, so we present the numbers of sentences, words and labels in Table~\ref{tab:probe-dataset}.

\paragraph{Training} We train classifiers for all probing tasks with an Adam optimizer at a starting learning rate of 5e-4 for 3,000 training steps with a batch size of 32, and evaluate on the development set every 50 training steps, halving the learning rate if no improvement is seen in 5 evaluation during training.

\end{document}